\documentclass{article}

\usepackage{arxiv}

\usepackage[utf8]{inputenc} % allow utf-8 input
\usepackage[T1]{fontenc}    % use 8-bit T1 fonts
\usepackage{hyperref}       % hyperlinks
\usepackage{url}            % simple URL typesetting
\usepackage{booktabs}       % professional-quality tables
\usepackage{amsfonts}       % blackboard math symbols
\usepackage{nicefrac}       % compact symbols for 1/2, etc.
\usepackage{microtype}      % microtypography
\usepackage{lipsum}		% Can be removed after putting your text content
\usepackage{graphicx}
\usepackage{natbib}
\usepackage{doi}

\usepackage{graphicx}
\usepackage{booktabs}
\usepackage{multirow}
\usepackage{floatrow}
\usepackage{amsmath}
\usepackage{cleveref}
\usepackage{bbding}
\usepackage{algorithm}
\usepackage{algorithmic}

\title{IceBench-S2S: A Benchmark of Deep Learning for Challenging Subseasonal-to-Seasonal Daily Arctic Sea Ice Forecasting in Deep Latent Space}

\author{Jingyi Xu$^{1,*}$, Shengnan Wang$^{1,*}$, Weidong Yang$^{1, }$\Envelope, Siwei Tu$^1$, \textbf{Lei Bai$^{2, }$\Envelope}, \textbf{Ben Fei$^{3, }$\Envelope}\\
	$^1$Fudan University, $^2$Shanghai AI Laboratory, $^2$Chinese University of Hong Kong\\
    \texttt{jyxu22@m.fudan.edu.cn}, 
    \texttt{wdyang@fudan.edu.cn},
    \texttt{baisanshi@gmail.com}, 
    \texttt{benfei@cuhk.edu.hk}\\
    $^{*}$ Equal Contributions, \Envelope Corresponding Authors
}

\date{}
\renewcommand{\shorttitle}{IceBench-S2S}

\begin{document}
\maketitle

\begin{abstract}
Arctic sea ice plays a critical role in regulating Earth's climate system, significantly influencing polar ecological stability and human activities in coastal regions.
Recent advances in artificial intelligence have facilitated the development of skillful pan-Arctic sea ice forecasting systems, where data-driven approaches showcase tremendous potential to outperform conventional physics-based numerical models in terms of accuracy, computational efficiency and forecasting lead times.
Despite the latest progress made by deep learning (DL) forecasting models, most of their skillful forecasting lead times are confined to daily subseasonal scale and monthly averaged values for up to six months, which drastically hinders their deployment for real-world applications, e.g., maritime routine planning for Arctic transportation and scientific investigation.
Extending daily forecasts from subseasonal to seasonal (S2S) scale is scientifically crucial for operational applications. 
To bridge the gap between the forecasting lead time of current DL models and the significant daily S2S scale, we introduce IceBench-S2S, the first comprehensive benchmark for evaluating DL approaches in mitigating the challenge of forecasting Arctic sea ice concentration in successive 180-day periods.
It proposes a generalized framework that first compresses spatial features of daily sea ice data into a deep latent space.
The temporally concatenated deep features are subsequently modeled by DL-based forecasting backbones to predict the sea ice variation at S2S scale.
IceBench-S2S provides a unified training and evaluation pipeline for different backbones, along with practical guidance for model selection in polar environmental monitoring tasks.
\end{abstract}

\section{Introduction}
%1. Importance and use of SIC (not SIT, and other Sea Ice Related variables.)
Arctic sea ice, as an essential component of the Earth system, exerts profound impacts both within and beyond the pan-Arctic region~\cite{Budikova2009role}. 
The coupled interactions between Arctic sea ice, atmosphere, and ocean influence the local weather, and the observed Arctic amplification is crucial for analyzing the global climate change~\cite{Serreze2011processes,rantanen2022arctic,zhou2024steady}. 
% Increasing Research Interest of DL (Related works, Combining Dynamical/Physical and DL models, and criticize altogether)
Researchers have established numerical~\cite{johnson2019seas5} and statistical models~\cite{wang2016predicting,wang2019subseasonal} to simulate the dynamical properties of sea ice for short-time forecasting.
However, their forecasting skills decline rapidly as forecasting leads gradually extend to the seasonal scale, primarily due to the lack of solid knowledge of physical processes~\cite{Merryfield2013multisystem} and potentially inaccurate initial and boundary conditions~\cite{zhang2022subseasonal}.
Recent studies on adopting deep learning (DL) approaches to forecast Arctic sea ice have demonstrated promising results while circumventing the expensive computational overheads of numerical models~\cite{schneider2023harnessing}.
For instance, DL-based methods have achieved skillful prediction of daily Arctic sea ice concentration (SIC) in 90 days~\cite{ren2023predicting} and prediction of monthly sea ice extent (SIE) for the subsequent two seasons compared to dynamical and statistical methods~\cite{andersson2021seasonal}.
Despite recent promising progress, the development of skillful DL models with daily forecasting lead times that transcend intra-seasonal scale is still under-explored. 
% and have shown prominent results. 
% For example, gaining more skillful prediction of daily Arctic sea ice concentration (SIC) in 90 days~\cite{ren2023predicting} and prediction of monthly sea ice extent (SIE) for the subsequent two seasons compared to dynamical and statistical methods~\cite{andersson2021seasonal}.

% Fig for annual mean SIC/SIE Trend
\begin{figure}
	\centering
	% \vspace{-1cm}
	\includegraphics[width=0.8\columnwidth]{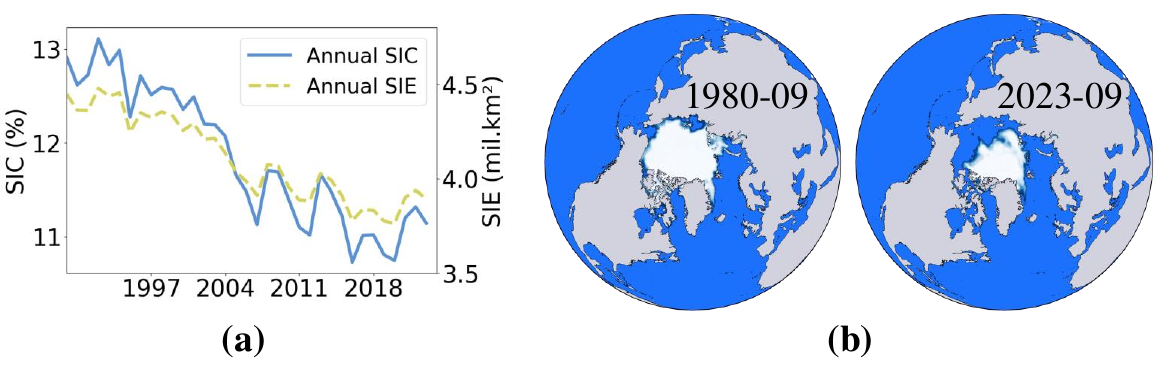} 
	% \vspace{-0.7cm}
	\caption{Variation of Arctic sea ice: (a) Annual trend of averaged Arctic sea ice concentration and sea ice extent over the last few decades. (b) Monthly average of sea ice concentration in September 1980 and 2023. }
	\label{fig:g02202}
	% \vspace{-0.2cm}
\end{figure}

%Although these models are relatively reliable for generating accurate predictions of daily sea ice concentration (SIC) at the sub-seasonal scale, their performances rapidly deteriorate as forecasting leads gradually extend to seasonal scale.
% 2. Why S2S, Economical, Environmental (Extreme Events), Develop coupled prediction systems
% Considering new opportunities for Arctic maritime transportation and scientific exploration brought by the consistent decline and negative trends of SIE during the melting season in the last few decades (as in Figure~\ref{fig:g02202}(a) and (b)), fulfilling accurate subseasonal-to-seasonal (S2S) forecasting is increasingly important for polar research~\cite{winton2022prospects} and uncovering connections between Arctic sea ice variation and extreme weather events in middle-to-low latitude~\cite{Liu2023s2sprediction}.
The consistent decline and negative trends of SIE during the melting season in the last few decades (as in Figure~\ref{fig:g02202}(a) and (b)) have opened new opportunities for Arctic maritime transportation and scientific exploration.
This trend highlights the growing importance of accurate subseasonal-to-seasonal (S2S) forecasting, both for polar research~\cite{winton2022prospects} and uncovering connections between Arctic sea ice variation and extreme weather events in middle-to-low latitudes~\cite{Liu2023s2sprediction}.
However, significant challenges remain in extending daily forecasting lead time to inter-seasonal scale and the spring prediction barrier (SPB) of Arctic SIC before the summer season\cite{bushuk2020mechanism}.
Overcoming these limitations could further advance the development of DL-based forecasting models~\cite{ren2024sicnet}.
Current DL approaches that perform at a seasonal scale mainly focus on the prediction of monthly average SIC, which inevitably fall short of forming an operational S2S Arctic sea ice forecasting system for at least two reasons:
% Besides, extending daily forecasting lead time to inter-seasonal scale and the spring prediction barrier (SPB) of Arctic SIC before summer season presents a greater challenge that could further advance the development of DL-based forecasting models~\cite{bushuk2020mechanism,ren2024sicnet}.  
% However, current DL approaches that perform at a seasonal scale mainly focus on the prediction of monthly average SIC, which inevitably fall short of forming an operational S2S Arctic sea ice forecasting system for at least two reasons:
(1) The averaged monthly SIC integrates the daily variation to a coarser temporal scale and therefore neglects the nature of the temporal and spatial continuum of weather and climate~\cite{Vitart2019subseasonal}.
(2) The captured monthly average pattern is useful for analyzing long-term trends, but the fine-grained information of the sea ice fluctuation is overlooked. Such absence of daily values on the inter-seasonal scale could be insufficient as a reference in real-world operations and therefore compromise the practicality of forecasting models.
% for reference in real-world applications.

%\textcolor{red}{
	%benchmark of statistical, dynamical models~\cite{bushuk2024predicting}
	%}

To facilitate the understanding of continual long-term daily variation of Arctic sea ice, advance the operational polar research, and bridge the gap between current DL-based sea ice models and significant S2S forecasting, we introduce IceBench-S2S - the first comprehensive benchmark of a DL approach to predict daily SIC for consecutive 180 days. 
%state-of-the-art DL time series forecasting models to predict daily SIC for consecutive 180 days. 
% Pose challenges and briefly explain methodology- LTF by TS models, Latent Space Compression and Related Climate Models, S2S in weather prediction
%\textcolor{red}{
	%Since directly extending the lead time of current DL models to a daily S2S forecasting scale would cause an excessive use of GPU memory %and increase the training cost, we propose to first spatially compress SIC data using an encoder and then model daily SIC variations in %latent space. 
	%% Quick summary of Tasks and stress on benefits
	%By adopting this paradigm, we could easily integrate various types of time series forecasting models and conduct experiments for fair %comparison.
	%}
Inspired by previous work~\cite{zheng2024spatio}, which utilizes multivariate empirical orthogonal functions (EOFs) to compress 2D sequential daily sea ice data into time series and leverages deep neural networks for embedding and prediction, IceBench-S2S proposes a generic DL framework, namely \underline{\textbf{S}}ea \underline{\textbf{I}}ce \underline{\textbf{F}}orecasting \underline{\textbf{E}}ngine. SIFE first compresses spatial daily sea ice data into a latent feature, then it employs DL time series models as the forecasting backbone to predict future values. By adopting this approach, IceBench-S2S could easily integrate various categories of time series DL models and comprehensively evaluate the forecasting skills of those models.

% Compress SIC into Latent space

% Motivation
%1. Subseasonal to Seasonal
% Mainly focus on daily, 6 average
%2. Challenging Task poses question: Can DL be the solution?
%(Limit of Statistical/Numerical Models) Capturing Long-term Trend and Variation

% Contribution
% Propose first benchmark on S2S sea ice forecasting based on DL approaches, Comprehensive Analysis
% Adopt latent space paradigm (compare to EOF), propose a challenge/scenario for TSF methods.

Our contributions are summarized as follows:
\begin{itemize}
	\item We propose the first benchmark, IceBench-S2S, that focuses on the evaluation of the DL approach for daily S2S forecasting of Arctic sea ice. 
	Thorough evaluations are conducted to investigate the potential of DL models to fulfill this challenging task.
	%This bridges the gap between DL-based time series forecasting approaches and
	\item Motivated by prior works, we propose a generic DL framework, SIFE, that compresses daily SIC data into a deep latent space to facilitate adopting and evaluating advanced DL-based time series forecasting models. 
	\item Our benchmark is mutually beneficial for AI and geoscience communities: the difficult S2S daily forecasting task integrated in IceBench-S2S poses significant challenges for future development of DL models;
	The newly constructed baselines of long-term daily SIC forecasting and our comprehensive analysis could further promote Arctic sea ice research and interdisciplinary development.
\end{itemize}

%Ref:
% Part I:
% Motivation Strong Evidence:
% [1]Role of Arctic sea ice in global atmospheric circulation: A review
% [2]Processes and impacts of Arctic amplification: A research synthesis
% [3]The Arctic has warmed nearly four times faster than the globe since 1979
% [4]Steady threefold Arctic amplification of externally forced warming masked by natural variability

%Part II:
% S2S for Arctic Sea Ice and Weather:
% [5]Subseasonal-to-seasonal prediction of arctic sea ice Using a Fully Coupled dynamical ensemble forecast system -- Hincast
% [6]Subseasonal-to-Seasonal Arctic Sea Ice Forecast Skill Improvement from Sea Ice Concentration Assimilation
% [7]Multi-system seasonal predictions of Arctic sea ice
% [8] The sub-seasonal to seasonal prediction project (S2S) and the prediction of extreme events
% [9] Progress in subseasonal to seasonal prediction through a joint weather and climate community effort
%[10] A machine learning model that outperforms conventional global subseasonal forecast models
% [11] Prospects for Seasonal Prediction of Summertime Trans-Arctic Sea Ice Path
% [12]Sub-Seasonal to Seasonal Prediction: The Gap Between Weather and Climate Forecasting 

\section{Related Works}

\textbf{Benchmarks in geoscience. }
% ClimaLearn, ClimaSet, ClimSim, ChaosBench
% In recent years, research on benchmarks in geoscience has emerged as a critical focus, primarily addressing two key aspects~\cite{kaltenborn2023climateset,nguyen2023climatelearn,yu2023climsim}. 
Recent years have seen growing emphasis on benchmarks in geoscience, primarily addressing two key aspects~\cite{kaltenborn2023climateset,nguyen2023climatelearn,yu2023climsim}. 
The first involves evaluating datasets (e.g., satellite products, meteorological variables) in terms of their accuracy and usability, which establishes regional validation standards~\cite{daly2006guidelines,loew2017validation}. 
For instance, systematic assessments of precipitation datasets like IMERG~\cite{tan2019imerg,huffman2015nasa} have provided essential benchmarks for refining satellite-based hydrological models~\cite{wang2017evaluation,pradhan2022review}. 
The second aspect focuses on performance comparisons of forecasting models through hyperparameter optimization, such as evaluating architectures like SmaAt-Unet~\cite{trebing2021smaat}, ConvLSTM~\cite{azad2019bi}, and DGMR~\cite{ravuri2021skilful}. 
These studies aim to identify optimal configurations for specific geoscientific tasks, such as extreme rainfall prediction~\cite{suri2024optimal}. 
Regarding sea ice benchmarks, IceBench~\cite{alkaee2025icebench} addresses the evaluation of the sea-ice type classification problem. For the forecasting benchmark, spatially damped anomaly persistence (SDAP)~\cite{niraula2021spatial}, which utilizes only sea ice concentration data, was proposed to evaluate dynamical models.
SDAP outperforms commonly used simple anomaly persistence and climatology baselines at subseasonal scales; its skillful seasonal forecasts compared to the best-performing dynamical forecast systems have made it a challenging benchmark method.
Considering the large variability of sea ice during melting seasons, the sea ice prediction community annually hosts the Sea Ice Outlook (SIO) contribution forum to gather sea ice predictions from worldwide institutes. A comprehensive multi-model benchmark~\cite{bushuk2024predicting} that exploits SIO prediction records was introduced to evaluate the skills of multiple dynamical and statistical models to forecast SIC and SIE in September.
However, benchmark studies that specifically focus on evaluating deep learning approaches for S2S daily forecasting remain relatively scarce.
%remain relatively scarce in scenarios involving high-dimensional geospatial variables, largely due to the limitations of traditional time-series models in capturing complex spatial-temporal interactions. 
In this work, we intend to bridge the gap and enrich the models in geoscience and sea ice forecasting.

\begin{figure*}[t]
    \centering
    \includegraphics[width=1.0\linewidth]{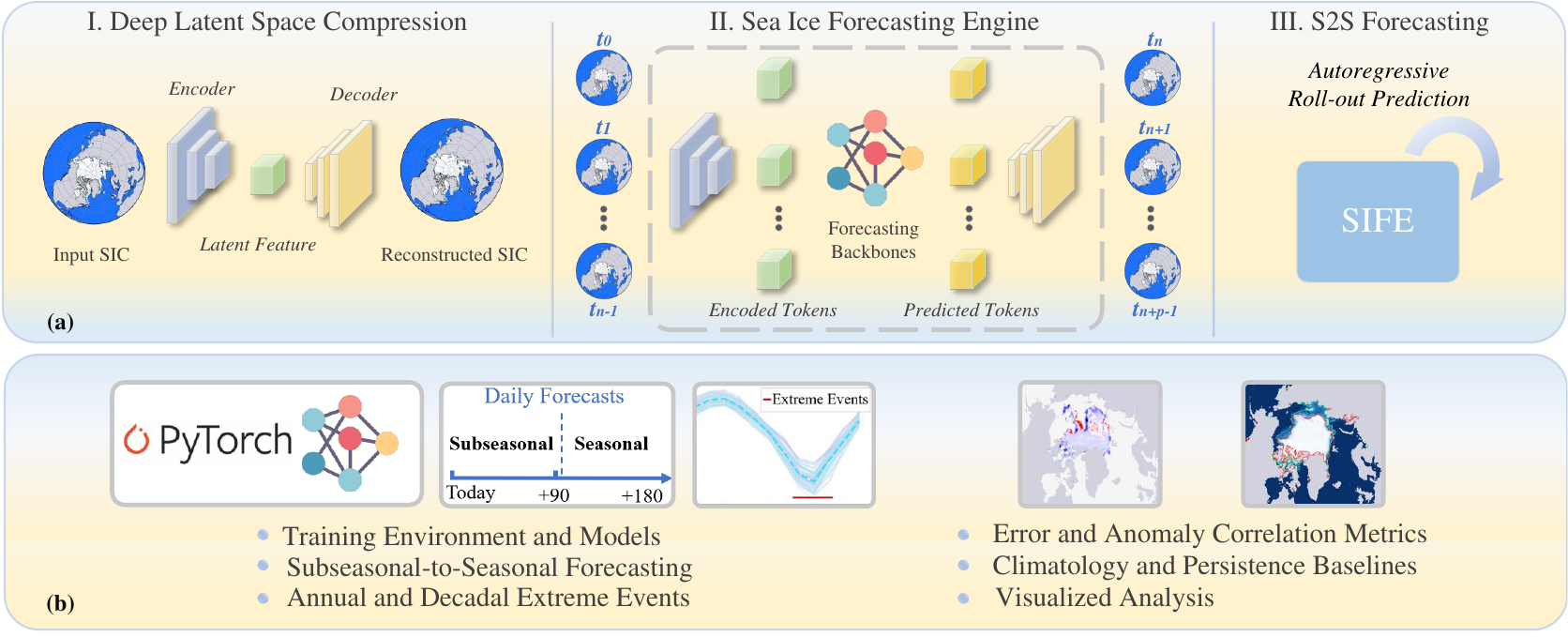}
     \caption{\textbf{IceBench-S2S}. \textbf{(a)} Forecasting Approach: \textit{Deep Latent Space Compression} for spatial representation, proposed generic framework \textit{Sea Ice Forecasting Engine} and autoregressive roll-out prediction for \textit{S2S Arctic Sea Ice Forecasting}, \textit{n} and \textit{p} represent the input and predicted time steps, respectively; \textbf{(b)} Key components include the model \textit{Training Environments}, challenging \textit{Tasks}, and comprehensive \textit{Evaluations}.}
    \label{fig:main}
    \vspace{-0.2cm}
\end{figure*}

\noindent\textbf{Deep learning for SIC forecasting. }
Researchers have proposed various approaches to forecasting SIC, encompassing numerical and statistical models~\cite{wang2013seasonal,yuan2016arctic}. 
However, numerical and statistical models usually rely on high-performance
computing of CPU clusters, which tends to result in complex debugging processes and uncertain parameterization.
Recently, deep learning models have drawn the attention of sea ice research communities and have been widely
investigated for Arctic sea ice forecasting~\cite{petrou2019prediction,kim2020prediction,ali2021sea,ali2022mt}. These methods utilize U-Net-based architectures to solve daily (SICNet~\cite{ren2022data}), or monthly average
(IceNet~\cite{andersson2021seasonal}, MT-IceNet~\cite{ali2022mt}) SIC forecasting. Although these U-Net-based architectures are built on top of LSTM~\cite{liu2021extended}, CNN~\cite{andersson2021seasonal}, or vision transformer, merging channels altogether could fail to fully exploit the temporal information inherent in sea ice modeling. Moreover, these methods, including the latest Transformer-based model~\cite{zheng2024spatio}, only concentrate on single temporal granularity SIC forecasting, and their daily forecasting leads are shorter than the seasonal scale.
%where the inter-granularity information from multi-granularity sea ice modeling is overlooked.

\section{IceBench-S2S}

\subsection{Sea Ice Concentration Dataset} % Component: Dataset
Considering the demonstrated effectiveness of leveraging SIC data to forecast future sea ice~\cite{niraula2021spatial,ren2022data}, we focus exclusively on SIC data for training, evaluation, and analysis in IceBench-S2S to set up S2S daily forecasting DL baselines, providing a foundation for future advancements.
In IceBench-S2S, we adopt the Climate Data Record of Passive Microwave Sea Ice Concentration G02202 Version 4 dataset~\cite{meier2021g02202} from the National Snow and Ice Data Center (NSIDC).
G02202 dataset records daily SIC data starting from October 25th, 1978, and provides the coverage of the pan-Arctic region (N:\(89.8^\circ\), S:\(31.1^\circ\), E:\(180^\circ\), W:\(-180^\circ\)). Each daily SIC data is formed of \(448 \times 304\) pixels and each pixel corresponds to the area of a \(25 \, \text{km} \times 25 \, \text{km}\) grid. 
The SIC data has a range of \(0\%\) to \(100\%\), and areas where the SIC value is greater than \(15\%\) indicate the SIE.
G02202 contains three subsets that are generated using different retrieval algorithms: Climate Data Record (CDR)~\cite{meier2021g02202}, NASA Team (NT)~\cite{cavalieri1984nasateam}, and Bootstrap (BT)~\cite{comiso1986bootstrap}. 
These algorithms are widely recognized in the field of sea ice concentration retrieval and are based on passive microwave remote sensing data.
We utilize the CDR-retrieved SIC dataset, which is the main dataset in G02202, to train and evaluate different models. 
The spatial resolution of the SIC data is maintained at a size of \(448 \times 304\).
The dataset covers the period up to June 30th, 2024, spanning a total of 16,686 days. In IceBench-S2S, it is divided into three subsets as follows: training period (1979–2015), validation period (2016–2019), and testing period (2020–2024).

\begin{table*}[t]\small
	% \begin{wraptable}{r}{0.8\textwidth}\small
		\centering
		\caption{Metrics utilized for IceBench evaluations.}
		%\label{tab:metrics}
		\resizebox{\linewidth}{!}{
			\begin{tabular}{c|c|c|c|c|c|c|c}
				\toprule
				\textbf{MSE}  & \textbf{RMSE}  & \textbf{MAE} & \textbf{NSE} & \textbf{ACC}  & \textbf{R$^2$} & \textbf{PSNR} & \textbf{SSIM} \\ \hline
				$ \frac{1}{n}\sum_{i=1}^n (y_i - \hat{y}_i)^2 $          &    $ \sqrt{\frac{1}{n}\sum_{i=1}^n (y_i - \hat{y}_i)^2} $                    & $  \frac{1}{n}\sum_{i=1}^n |y_i - \hat{y}_i| $                                     & $ 1 - \frac{\sum_{i=1}^n (y_i - \hat{y}_i)^2}{\sum_{i=1}^n (y_i - \bar{y})^2} $  & 
				$ \frac{\frac{1}{n}\sum_{i=1}^n (\hat{y}_i - C) (y_i - C)}{\sqrt{\frac{1}{n}\sum_{i=1}^n (\hat{y}_i - C)^2}\sqrt{\frac{1}{n}\sum_{i=1}^n (y_i - C)^2}} $  & \multicolumn{1}{c|}{$1 - \frac{\sum (y_i - \hat{y}_i)^2}{\sum (y_i - \bar{y})^2}$} & \multicolumn{1}{c|}{$10\log_{10}\left(\frac{\mathrm{MAX}^2}{\mathrm{MSE}}\right)$} & \multicolumn{1}{c}{$\frac{(2\mu_x\mu_y + C_1)(2\sigma_{xy} + C_2)}{(\mu_x^2 + \mu_y^2 + C_1)(\sigma_x^2 + \sigma_y^2 + C_2)}$} \\ \bottomrule
			\end{tabular}
		}
            %\caption{Metrics utilized for IceBench evaluations.}
		\label{tab:metrics}
		% \end{wraptable}
\end{table*}

\subsection{Task Overview} % Component: Benchmarks
\label{subsec:task}

\noindent\textbf{Subseasonal-to-Seasonal daily forecasting.} 
The primary task of IceBench-S2S is to evaluate daily Arctic SIC forecasting skills on a seasonal scale, i.e., forecasting SIC for the next 180 days.
%addresses two fundamental challenges in Arctic SIC forecasting:  
Considering the dimensions of daily data and forecasting future SIC requires modeling of sequences, the S2S task poses two fundamental challenges for deep learning approaches to address:
(1) \textit{High-Dimensional Data Compression}: Effective representation of daily SIC grids that project sea ice data into deep latent space while preserving critical spatial patterns;
(2) \textit{Temporal Sequence Modeling}: Extract significant signals inherent in historical SIC data to accurately predict future daily values.
%Consistent evaluation of temporal models across multiple prediction horizons.  
% Latent Space Benefits ：
% Compression for Easy access and modification of forecasting backbones
% Unified Encoder-Decoder for fair comparison
%The task formalization comprises:  
We formalize the above challenges as follows: 
\begin{equation}
\begin{aligned}
&\text{Compression:} \min_{f,g} \mathbb{E}_t[\|\mathbf{X}_t - g(f(\mathbf{X}_t))\|^2], \quad \\
&\text{Forecasting:} \hat{\mathbf{z}}_{t+\Delta} = \phi_{\text{TS}}(\mathbf{z}_{t-k:t}) \quad \Delta \in \{7,15,30,180\},
\end{aligned}
\end{equation}
%&\text{where } \gamma = \tfrac{448\times304}{\dim(f(\mathbf{X}_t))} \gg 1, \\
where $\mathbf{X}_t$ is daily SIC data, encoder $f$ compress input data and decoder $g$ restores SIC from compressed latent representation. The compression ratio $\gamma = \tfrac{448\times304}{\dim(f(\mathbf{X}_t))} \gg 1$. $\mathbf{z}$ is the latent feature,  $\mathbf{z}_{t-k:t}$ are input sequence of features and $\hat{\mathbf{z}}_{t+\Delta}$ are prediction results. $\phi_{\text{TS}}$ stands for the backbone of DL-based time series models. $\Delta$ is customizable for the forecasting lead time, i.e., we benchmark forecasting skills of a model in the next 7, 15, 30, and 180 days. 
The key requirements for DL models are as follows: (1) Maintain over 90\% spatial correlation in reconstructions; 
(2) Ensure identical hyperparameter spaces across $\phi_{\text{TS}}$ implementations; 
(3) Evaluate temporal models through standardized protocols and commonly used metrics after spatial decoding.

\noindent\textbf{Annual and decadal extreme events.} 
The annual minimum of sea ice extent usually occurs in September, which is the end of the melting season. 
Accurately forecasting these annual extremes poses an additional challenge that requires models to predict both temporally and numerically align with the ground truth. 
Based on the analysis of trends in Figure~\ref{fig:g02202}(a), we pick the historical minimum SIE in September 2016 as the decadal extreme event, corresponding to the lowest annual mean of SIC since 1979. 
The minimum SIE in September 2022 is selected as the annual extreme event during the test set, since the annual mean SIC has drastically increased compared to previous years, indicating a potential shift in climatology.

% \textbf{Hindcast.} 
% Additionally, we analyze hindcast (or reforecast) of annual extremes of SIE in the range of the training dataset to further validate the consistency of models' predictive skill. 
% We attach detailed analysis and plots to the Appendix.

% AE Latent Space Compression
%\begin{table}[ht]
%% \begin{wraptable}{r}{0.6\textwidth}\small
%% \centering
%% \vspace{-1cm}
%\caption{Compression quality of different retrieval algorithm and latent space dimensions.}
%\label{tab:sic_compress}
%\begin{tabular}{c|c|c|c|c|c}
%\toprule
%\textbf{Dataset}   & \textbf{MSE}$\downarrow$ & \textbf{MAE}$\downarrow$ & \textbf{PSNR}$\uparrow$ & \textbf{SSIM}$\uparrow$ & \textbf{NSE}$\uparrow$ \\ \hline
% CDR$_{4096}$ & \textbf{0.0029} & \textbf{0.0123} & \textbf{25.35} & \textbf{0.9243} & \textbf{0.9705}\\ 
% CDR$_{1024}$ & 0.0031 & 0.0125 & 25.12 & 0.9211 & 0.9689     \\ \hline
% NT$_{4096}$  & \textbf{0.0023} & \textbf{0.0137} & \textbf{26.45} & \textbf{0.9048} & \textbf{0.9732} \\ 
% NT$_{1024}$  & 0.0030 & 0.0163 & 25.31 & 0.8810 & 0.9660 \\ \hline
% BT$_{4096}$  & \textbf{0.0026} & \textbf{0.0126} & \textbf{25.96} & \textbf{0.9177} & \textbf{0.9752} \\ 
% BT$_{1024}$  & 0.0035 & 0.0148 & 24.55 & 0.9081 & 0.9650 \\ \bottomrule
%\end{tabular}
%\end{table}
%    
%% \end{wraptable}

\begin{table}[t]\small
	% \begin{wraptable}{r}{0.6\textwidth}\small
		\centering
		% \vspace{-1cm}
		\caption{Compression quality of different retrieval algorithms and latent space dimensions. CDR$_{L}$, NT$_{L}$, and BT$_{L}$ represent that the dimension of compressed deep latent space is 4096, while the dimension of CDR$_{S}$, NT$_{S}$, and BT$_{S}$ is equal to 1024. The quality improvement of quadrupling the latent dimension is marginal.}
		\begin{tabular}{c|c|c|c|c|c}
			\toprule
			\textbf{Dataset}   & \textbf{MSE}$\downarrow$ & \textbf{MAE}$\downarrow$ & \textbf{PSNR}$\uparrow$ & \textbf{SSIM}$\uparrow$ & \textbf{NSE}$\uparrow$ \\ \hline
			 CDR$_{L}$ & \textbf{0.0029} & \textbf{0.0123} & \textbf{25.35} & \textbf{0.9243} & \textbf{0.9705}\\ 
			 CDR$_{S}$ & 0.0031 & 0.0125 & 25.12 & 0.9211 & 0.9689     \\ \hline
			 NT$_{L}$  & \textbf{0.0023} & \textbf{0.0137} & \textbf{26.45} & \textbf{0.9048} & \textbf{0.9732} \\ 
			 NT$_{S}$  & 0.0030 & 0.0163 & 25.31 & 0.8810 & 0.9660 \\ \hline
			 BT$_{L}$  & \textbf{0.0026} & \textbf{0.0126} & \textbf{25.96} & \textbf{0.9177} & \textbf{0.9752} \\ 
			 BT$_{S}$  & 0.0035 & 0.0148 & 24.55 & 0.9081 & 0.9650 \\  \bottomrule
		\end{tabular}
		%\caption{Compression quality of different retrieval algorithms and latent space dimensions. CDR$_{L}$, NT$_{L}$, and BT$_{L}$ represent that the dimension of compressed deep latent space is 4096, while the dimension of CDR$_{S}$, NT$_{S}$, and BT$_{S}$ is equal to 1024. The quality improvement of quadrupling the latent dimension is marginal.}
		\label{tab:sic_compress}
	\end{table}
	
	% \end{wraptable}

% Latent Space Forecasting 

\subsection{Sea Ice Forecasting Engine}  % Component: DL Models
\label{subsec:Methodology}
%Our framework directly addresses the S2S forecasting requirements from Section~\ref{subsec:task}  through:  
To bridge the gap between the under-explored DL-based S2S daily sea ice forecasting task, we propose the sea ice forecasting engine, \textbf{SIFE}), a generic DL framework that directly addresses the S2S daily forecasting challenges in the previous section through:
(1) \textit{Spatial Compression}: A Swin Transformer-based~\cite{liu2022swin} autoencoder (Swin-AE) for spatial dimension reduction while preserving geophysical patterns;
%Temporal Benchmarking
(2) \textit{Temporal Forecasting}: Unified experimental protocols for fair comparison of various sequence forecasting backbones across different horizons, covering from subseasonal 7 days to inter-seasonal 180 days. 

\noindent\textbf{Deep latent space representation.} 
The Swin-AE architecture, as in Figure~\ref{fig:main}(a).I, compresses SIC grids through hierarchical encoding and latent projection. Using Swin Transformer blocks with shifted window attention, the encoder $f$ processes $448\!\times\!304$ inputs via:
%$f_\theta$
\begin{equation}
\begin{aligned}
&\mathbf{X}^{(0)}_{t} = \text{Conv2D}_{2\times2}(\mathbf{X}_t), \\
&\mathbf{X}^{(l)}_{t} = \text{SwinBlock}^{(l)}(\mathbf{X}^{(l-1)}_{t}) \quad (l=1,...,4),
\end{aligned}
\end{equation}
where $\mathbf{X}^{(0)}_{t} \in \mathbb{R}^{224\times152\times8} $, encoder
progressively downsampling spatial dimensions by $2\times$ per stage while doubling channels. 
The compressed latent code $\mathbf{z}_t \in \mathbb{R}^{1024}$ is obtained through:
\begin{equation}
\mathbf{z}_t = \text{MLP}(\text{Flatten}(\mathbf{X}^{(4)}_{t})).
\end{equation}
%$g_\phi$
The decoder $g$ mirrors encoder operations with upsampling blocks, reconstructing $\tilde{\mathbf{X}}_t$ from $\mathbf{z}_t$ through inverse spatial transformations. 
This design preserves spatial patterns with a compression ratio $\gamma = \frac{448\times304}{1024} = 133$. 

%Sea ice forecasting.
\noindent\textbf{Forecasting backbones.} The compressed latent representation $\{\mathbf{z}_t\}$ are fed into temporal models to predict future variations.
The collection of temporal model $\mathcal{M}_{\text{TS}}$ in IceBench-S2S encompasses four architectural paradigms: 
%\begin{equation}\small
%\mathcal{M}_{\text{TS}} = \left\{
%\begin{aligned}
%&\text{Transformer}, \text{iTransformer}, \text{Informer}, \text{PatchTST} && \text{(Attention-based)} \\
%&\text{DLinear}, \text{NLinear} && \text{(Linear Baselines)} \\
%&\text{TimeMixer}, \text{SCInet} && \text{(Multi-scale Decomposition \& Hybrid)} \\
%&\text{CycleNet} && \text{(Periodic Architectures)}
%\end{aligned}
%\right.
%\end{equation}
\begin{equation}\small
	\mathcal{M}_{\text{TS}} = \left\{
	\begin{aligned}
		&\textbf{a}:\text{Transformer}, \text{iTransformer}, \text{Informer}, \text{PatchTST} && \\
		&\textbf{b}:\text{DLinear}, \text{NLinear} &&  \\
		&\textbf{c}:\text{TimeMixer}, \text{SCInet} &&  \\
		&\textbf{d}:\text{CycleNet} 
	\end{aligned}
	\right.
\end{equation}
\textbf{\textit{a. Attention-based}}: Transformer~\cite{vaswani2017attention}, iTransformer~\cite{liu2023itransformer}, Informer~\cite{zhou2021informer} and PatchTST~\cite{nie2023patchtst} are attention-based models. 
They tokenize time series data and utilize a multi-head attention mechanism for modeling sequential dependencies in the feature space.
\textbf{\textit{b. Linear Baselines}}: DLinear and NLinear~\cite{zeng2022aretransformer} are essentially linear models. 
The former employs trend and seasonality decomposition and separately models those components for improved forecasting skill. 
The latter is more focused on adopting a normalization method to mitigate the distribution shift that is commonly observed in real-world data.
\textbf{\textit{c. Multi-scale Decomposition \& Hybrid}}: TimeMixer~\cite{ekambaram2023tsmixer} and SCINet~\cite{liu2022scinet} decompose time series by downscaling input data into subsequences and leveraging their interdependency to facilitate prediction. 
\textbf{\textit{d. Periodic Architectures}}: CycleNet~\cite{lin2024cyclenet} utilizes the discrete Fourier transform, a classic decomposition approach in signal processing, to explicitly identify cyclic components that could promote the performance.
All models are trained under identical experimental conditions to ensure a fair comparison. 
They learn patterns and temporal dependencies in the compressed data to make predictions about future sea-ice states. 
The selection of these time series models aims to cover a wide range of approaches for predicting SIC in the compressed deep latent space and to establish a comprehensive benchmark for S2S forecasting.

\begin{table}[t]\small
	% \begin{wraptable}{r}{0.7\textwidth}\small
		% \vspace{-1cm}
		\centering
		\caption{Robustness of learned deep latent representation. The spatial autoencoder trained on the CDR dataset performs sufficiently well on both NT and BT datasets without the need for fine-tuning.}
		\begin{tabular}{c|c|c|c|c|c}
			\toprule
			\textbf{Dataset}   & \textbf{MSE}$\downarrow$ & \textbf{MAE}$\downarrow$ & \textbf{PSNR}$\uparrow$ & \textbf{SSIM}$\uparrow$ & \textbf{NSE}$\uparrow$ \\ \hline
			CDR$_{L}$-NT &\textbf{0.003} &\textbf{0.014} &24.70 &\textbf{0.914} &\textbf{0.966}\\ 
			 CDR$_{S}$-NT & \textbf{0.003} & 0.016 & \textbf{24.73} & 0.902 & 0.960     \\ \hline
			CDR$_{L}$-BT  &\textbf{0.003} &0.016 &\textbf{24.88} &0.906 &0.961 \\ 
			 CDR$_{S}$-BT  & 0.004 & \textbf{0.014} & 24.25 & \textbf{0.909} & \textbf{0.962} \\ \bottomrule
		\end{tabular}
		%\caption{Robustness of learned deep latent representation. The spatial autoencoder trained on the CDR dataset performs sufficiently well on both NT and BT datasets without the need for fine-tuning.}
		\label{tab:sic_compress_roubustness}
	\end{table}

\begin{table*}[t]\tiny
	\centering
	\caption{Performance of SIFE with different time series forecasting backbones and the proposed baseline on the CDR dataset.}
	%\label{tab:ts_forecasting}
	\resizebox{\linewidth}{!}{
		\begin{tabular}{l|cccccc|cccccc|cccccc} \toprule
			\hline
			& \multicolumn{6}{c|}{\textbf{7 Days}}                       & \multicolumn{6}{c|}{\textbf{6 Months' Average}}                     & \multicolumn{6}{c}{\textbf{S2S (180 Days})}                         \\ \cline{2-19} 
			\multirow{-2}{*}{\textbf{Methods}} & MAE $\downarrow$    & $R^{2}$ $\uparrow$ & NSE $\uparrow$    & MSE $\downarrow$ & RMSE $\downarrow$ &  ACC $\uparrow$  & MAE $\downarrow$   & $R^{2}$ $\uparrow$ & NSE $\uparrow$    & MSE $\downarrow$ & RMSE $\downarrow$ &  ACC $\uparrow$  & MAE $\downarrow$   & $R^{2}$ $\uparrow$ & NSE $\uparrow$   & MSE $\downarrow$ & RMSE $\downarrow$ & ACC $\uparrow$   \\ \hline
			
			Transformer~\cite{vaswani2017attention}         & 0.0163 & 0.9215 & 0.9134 & 0.0064 & 0.0793 & 0.8165          & 0.0161 & 0.9324  & 0.9238 & 0.0064 & 0.0798 & 0.8154            & 0.0289 & 0.8298 & 0.8098 & 0.0159 & 0.1243 & 0.5405 \\
			
			iTransformer~\cite{liu2023itransformer}       & 0.0151 & 0.9397  & 0.9332 & 0.0051 & 0.0706 & 0.8544       & 0.0169 & 0.9353 & 0.9271 & 0.0061 & 0.0778 & \underline{0.8216}             & 0.0491 & 0.6621 & 0.6239 & 0.0303 & 0.1693 & 0.0472  \\
			
			Informer~\cite{zhou2021informer}            & 0.0164 & 0.9253 & 0.9174 & 0.0061 & 0.0777 & 0.8202         & 0.0168 & \textbf{0.9416} & \textbf{0.9340}  & \underline{0.0060}  & \underline{0.0771} & 0.8193       	& 0.0265 & 0.846 & 0.8283 & 0.0140 & 0.1160 & 0.6156  \\
			
			PatchTST~\cite{nie2023patchtst}           & 0.0128 & 0.9476 & 0.9419 & 0.0044 & 0.0661 & 0.8691          & 0.0169 & 0.9253  & 0.9157 & 0.0075 & 0.0864 & 0.7945       & 0.0224 & 0.8833 & 0.8692 & 0.0112 & 0.1051 & 0.6953 \\
			
			TimeMixer~\cite{ekambaram2023tsmixer}          & 0.0141 & 0.9443 & 0.9382 & 0.0047 & 0.0682 & 0.8561			& 0.0196 & 0.9232   & 0.9135 & 0.0072& 0.0846 & 0.7896         & 0.0365 & 0.7401 & 0.7108 & 0.0232 & 0.1480 & 0.3121 \\
			
			SCINet~\cite{liu2022scinet}             & 0.0161 & 0.9246 & 0.9165 & 0.0063 & 0.0787 & 0.8058           & \underline{0.0155} & 0.9279  & 0.9188 & 0.0072 & 0.0845 & 0.7909         & \textbf{0.0170}  & \textbf{0.9212} & \textbf{0.9116} & 0.0076 & 0.0867 & 0.8087 \\
			
			DLinear~\cite{zeng2022aretransformer}            & 0.0155 & 0.9346 & 0.9274 & 0.0055 & 0.0737 & 0.8289         & 0.0165 & 0.9317 & 0.9231 & 0.0068 & 0.0820 & 0.8194  			& 0.0195 & 0.9159 & 0.9058 & 0.008 & 0.0886 & 0.7954  \\
			
			NLinear~\cite{zeng2022aretransformer}            & 0.0164 & 0.9185 & 0.9108 & 0.0058 & 0.0764 & 0.8164  		& 0.0178 & 0.9239  & 0.9155 & 0.0074 & 0.0859 & 0.8080            & 0.0298 & 0.8074 & 0.7838 & 0.0186 & 0.1311 & 0.5451 \\
			
			CycleNet~\cite{lin2024cyclenet}           & 0.0143 & 0.9448 & 0.9385 & 0.0049 & 0.0691 & 0.8519            & 0.0204 & 0.9312 & 0.9224 & 0.0065 & 0.0807 & 0.8185  			& 0.0511 & 0.6172 & 0.5724 & 0.0355 & 0.1842 & 0.3752\\
			
			\textbf{SIFE-Ensemble S2S Baseline}           & - & - & - & - & - & -               & - & - & - & - & - & -               & \underline{0.0175} & \underline{0.9191}  & \underline{0.9111} & \textbf{0.0061}  & \textbf{0.0776} & \textbf{0.8155} \\
			\hline
			
			SDAP~\cite{niraula2021spatial}           & \textbf{0.0071} & \textbf{0.9803} & \textbf{0.9780} & \textbf{0.0018} & \textbf{0.0424} & \textbf{0.9431}           & \textbf{0.0152} & \underline{0.9400} & \underline{0.9331} & \textbf{0.0053} & \textbf{0.0725} & \textbf{0.8318} 			& 0.0230 & 0.8939 & 0.8805 & 0.106 & 0.1028 & 0.7320  \\
			
			% \rowcolor[HTML]{B4D7E2}
			%\rowcolor{lightgrey!20}
			Persistence        & \underline{0.0072} & \underline{0.9672} & \underline{0.9672} & \underline{0.0025} & \underline{0.0494} & \underline{0.9269}  			& 0.0536 & 0.5735 & 0.5174 & 0.0441 & 0.2049 &  0.0131				& 0.0506 & 0.5830 & 0.5307 & 0.0413 & 0.1994 & 0.0134  \\
			% \rowcolor[HTML]{0077BE}
			%\rowcolor{oceanicblue!20}
			Climatology        & 0.0248  & 0.8641  & 0.8492 & 0.0117 & 0.1074 & 0.6638         & 0.0247 & 0.8643 & 0.8493 & 0.0118 & 0.1081 & 0.6733 				& 0.0247 & 0.8726 & 0.8676 & 0.118 & 0.1085 & 0.6836  \\ \hline
		\bottomrule
		\end{tabular}
	}
	%\caption{Performance of SIFE with different time series forecasting backbones and the proposed baseline on the CDR dataset.}
	\label{tab:ts_forecasting}
\end{table*}

\begin{figure*}[t]
    \centering
    \includegraphics[width=1.0\linewidth]{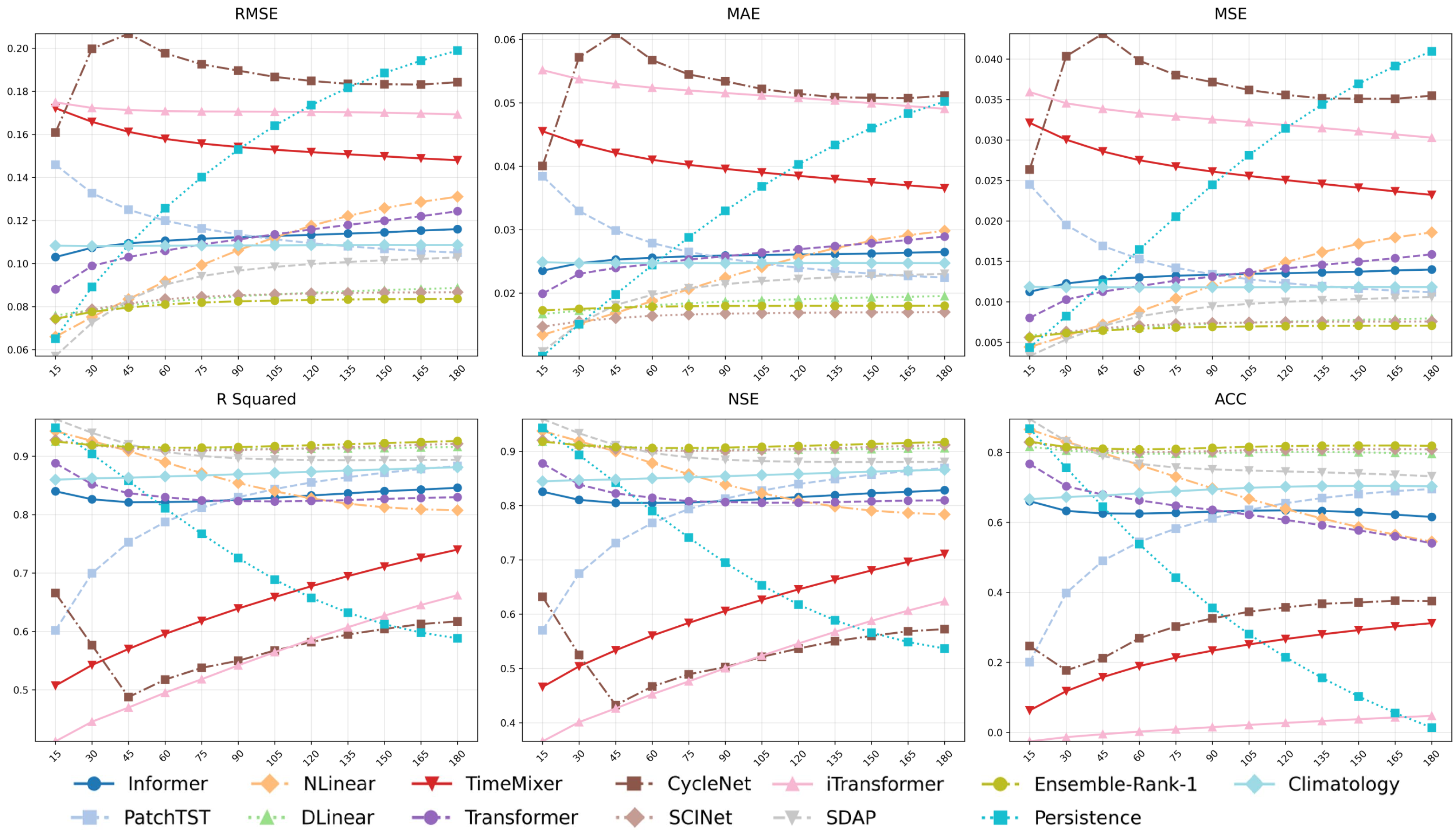}
    % \vspace{-0.3cm}
    \caption{\textbf{Main results of S2S forecasting.} DL models along with conventional baselines, which are commonly used for validating numerical and statistical models, are evaluated under metrics outlined in Table~\ref{tab:metrics} on the test dataset. Most of the DL models suffer from predictive skill loss once the forecasting lead time surpasses one month. }
    \label{fig:exp_overall}
\end{figure*}

\noindent\textbf{Auroregressive roll-out prediction and SIFE ensembles.}
To achieve the consecutive 180-day forecasting lead, we adopt a rolling-based training approach that adapts time series forecasting backbones $\mathcal{M}_{\text{TS}}$ to the S2S forecasting scenario. Considering the rolling window longer than 15 days would cause an out-of-memory issue during training of some backbones, we set the input and output length as 15 to ensure all backbones could be properly trained, i.e., SIFE autoregressively takes the 15-day prediction of its own as the input, and repeats the rolling stage 12 times to generate a total of 180 SIC forecasts. We treat the 15-day autoregressive SIFE as a recurrent neural network and leverage the teacher forcing strategy with a gradually decaying forcing ratio for a more stable training over the S2S time scale. Moreover, the model ensemble approach is exploited to form an S2S daily forecasting DL baseline.

\noindent{\textbf{Implementation.}}
We implement the benchmark under a unified Python framework for impartial comparison and consistency. 
The PyTorch package is used for building deep neural networks with cuDNN-v9.1 and CUDA-v11.7 as
the computational back-end. We are using the default setting of each forecasting backbone from the original paper. Hyperparameter settings and training details are attached to the Appendix. All experiments are performed on the same server with an NVIDIA A100 80GB GPU.

\subsection{Evaluation Protocol} % Component: Evaluation
\noindent\textbf{Research questions.}
We perform IceBench-S2S to explore the following:
(\textbf{RQ1}) Is the proposed deep latent space representation in the SIFE framework sufficient for representing the spatial distribution of Arctic sea ice?
% Response Exp: Compression, Robustness
(\textbf{RQ2}) How does the proposed SIFE compare to the previous benchmark in terms of evaluation metrics on daily, seasonal, and S2S scales?
% Response Exp: 7 Days, 6 Months, S2S
(\textbf{RQ3}) Could SIFE mitigate the  Spring Prediction Barrier based solely on the SIC data and accurately forecast the September SIE?
% Response Exp

\noindent\textbf{Metrics.} The reconstruction quality is evaluated using five key metrics: pixel-wise accuracy, mean squared error (MSE), mean absolute error (MAE), the information-theoretic fidelity peak signal-to-noise ratio (PSNR), structural similarity (SSIM), and hydrological modeling capability Nash-Sutcliffe efficiency (NSE)~\cite{NASH1970282}. 
These metrics collectively assess the fidelity of the reconstructed images. 
For the evaluation of forecasting skills, the results are assessed using MSE, MAE, the coefficient of determination (R²), the anomaly correlation coefficient (ACC), and NSE, which provide a comprehensive overview of the models' predictive performance. 
All these metrics are formally defined in Table~\ref{tab:metrics}, where $y$ is the ground truth SIC, $\bar{y}$ is the spatially averaged ground truth, $\hat{y}$ is the predicted value, and $C$ in ACC stands for the calculated climatology.

\noindent\textbf{Statistical and dynamical baselines.} The commonly used persistence and climatology forecasts are calculated as simple baselines. SDAP~\cite{niraula2021spatial} is implemented to represent the performance of dynamical sea ice forecasting systems. Moreover, we leverage SIO-reported predictions as a competitive baseline for evaluating the forecasting skill in the critical and challenging September.

\section{Benchmark Results and Analysis}

%%% One Row Version
% \usepackage{multirow}
\begin{table*}[t]\small
	\centering
    \caption{Performance of proposed SIFE under Sea Ice Outlook settings. Both fixed-parameter and fine-tuned SIFE demonstrate competitive forecast skill in the early spring (90 and 120 days lead time before September 1$^{st}$, indicating their potential for mitigating the challenge of spring prediction barrier by only using SIC data. }
	\resizebox{\linewidth}{!}{
		\begin{tabular}{l|ccc|ccc|ccc|ccc} \toprule
			\hline
			& \multicolumn{3}{c|}{\textbf{120 Days Lead Time}}                       & \multicolumn{3}{c|}{\textbf{90 Days Lead Time}}                     & \multicolumn{3}{c}{\textbf{60 Days Lead Time}}  & \multicolumn{3}{c}{\textbf{30 Days Lead Time}}                       \\ \cline{2-13} 
			\multirow{-2}{*}{\textbf{Methods}} & RMSE (Detrend) $\downarrow$    & ACC (Detrend) $\uparrow$ & ACC $\uparrow$    & RMSE (Detrend) $\downarrow$    & ACC (Detrend) $\uparrow$ & ACC $\uparrow$   & RMSE (Detrend) $\downarrow$    & ACC (Detrend) $\uparrow$ & ACC $\uparrow$   & RMSE (Detrend) $\downarrow$    & ACC (Detrend) $\uparrow$ & ACC $\uparrow$   \\ \hline
			
			\textbf{Sea Ice Outlook (Sea Ice Extent)}         & 0.4425 & 0.5758 & 0.8096       & 0.3463 & \textbf{0.8061} & \textbf{0.8895}          & 0.2937 & \textbf{0.8774}  & \textbf{0.9207}       & \textbf{0.1682} & \textbf{0.9710} & \textbf{0.9733}  \\

			SIFE-S2S-Median (Fixed)        & 0.6030 & 0.5136 & 0.7774       & 0.6657 & 0.5173 & 0.7803          & 0.5923 & 0.5032  & 0.7695       & 0.6351 & 0.5058 & 0.7181  \\
			
			SIFE-Ensemble-Rank-1 (Fixed)         & 0.4089 & 0.6571 & 0.7960       & 0.3889 & \underline{0.7790} & 0.8225          & 0.4050 & 0.7468  & 0.8015       & 0.4826 & 0.6119 & 0.8132  \\
			
			SIFE-S2S-Best (Fixed)        & \underline{0.4036} & \underline{0.6681} & \underline{0.8128}       & 0.4182 & 0.7662 & 0.8143          & 0.4274 & \underline{0.7608}  & 0.8187       & 0.4879 & 0.5561 & 0.7987  \\									

			SIFE-S2S-Best (Finetune)         & \textbf{0.2654} & \textbf{0.6866} & \textbf{0.8282}       & \textbf{0.2673} & 0.6901 & \underline{0.8329}          & \textbf{0.2689} & 0.6922  & \underline{0.8297}       & \underline{0.2678} & \underline{0.6892} & \underline{0.8260}  \\				
			\hline
			\hline
			
			\textbf{Sea Ice Outlook (Sea Ice Concentration)}         & \textbf{0.2355}& 0.4310 & 0.4799       & \textbf{0.2282}& 0.4343 & 0.4971          & \textbf{0.2074}& 0.4943  & 0.5563       & \textbf{0.1518}& 0.6753 & 0.7494  \\

			SIFE-S2S-Median (Fixed)         & 0.3869 & 0.5788 & 0.7891       & 0.3646 & 0.6109 & \underline{0.8062}          & 0.3445 & 0.6483  & \textbf{0.8275}       & 0.3366 & 0.6668 & 0.8270 \\
			
			SIFE-Ensemble-Rank-1 (Fixed)         & 0.3294 & \underline{0.6686} & \underline{0.7943}       & 0.3148 & \underline{0.6700} & 0.7935          & 0.2980 & \underline{0.6762}  & 0.7989       & 0.2699 & 0.7381 & \textbf{0.8398}  \\
			
			SIFE-S2S-Best (Fixed)         & 0.3333 & 0.6644 & 0.7824       & 0.3418 & 0.6519 & 0.7720          & 0.3441 & 0.6508  & 0.7689       & 0.2879 & \underline{0.7434} & 0.8282  \\									
			
			SIFE-S2S-Best (Finetune)         & \underline{0.2555} & \textbf{0.7531} & \textbf{0.8248}       & \underline{0.2540} & \textbf{0.7534} & \textbf{0.8233}          & \underline{0.2556} & \textbf{0.7571}  & \underline{0.8255}       & \underline{0.2603} & \textbf{0.7588} & 0.8250  \\	 \hline

		\bottomrule
		\end{tabular}
	}
	%\caption{Performance of proposed SIFE under Sea Ice Outlook settings. Both fixed-parameter and fine-tuned SIFE demonstrate competitive forecast skill in the early spring (90 and 120 days lead time before September 1$^{st}$, indicating their potential for mitigating the challenge of spring prediction barrier by only using SIC data. }
	\label{tab:sio_forecasting}
\end{table*}

\begin{figure*}[t]
    \centering
    \includegraphics[width=1.0\linewidth]{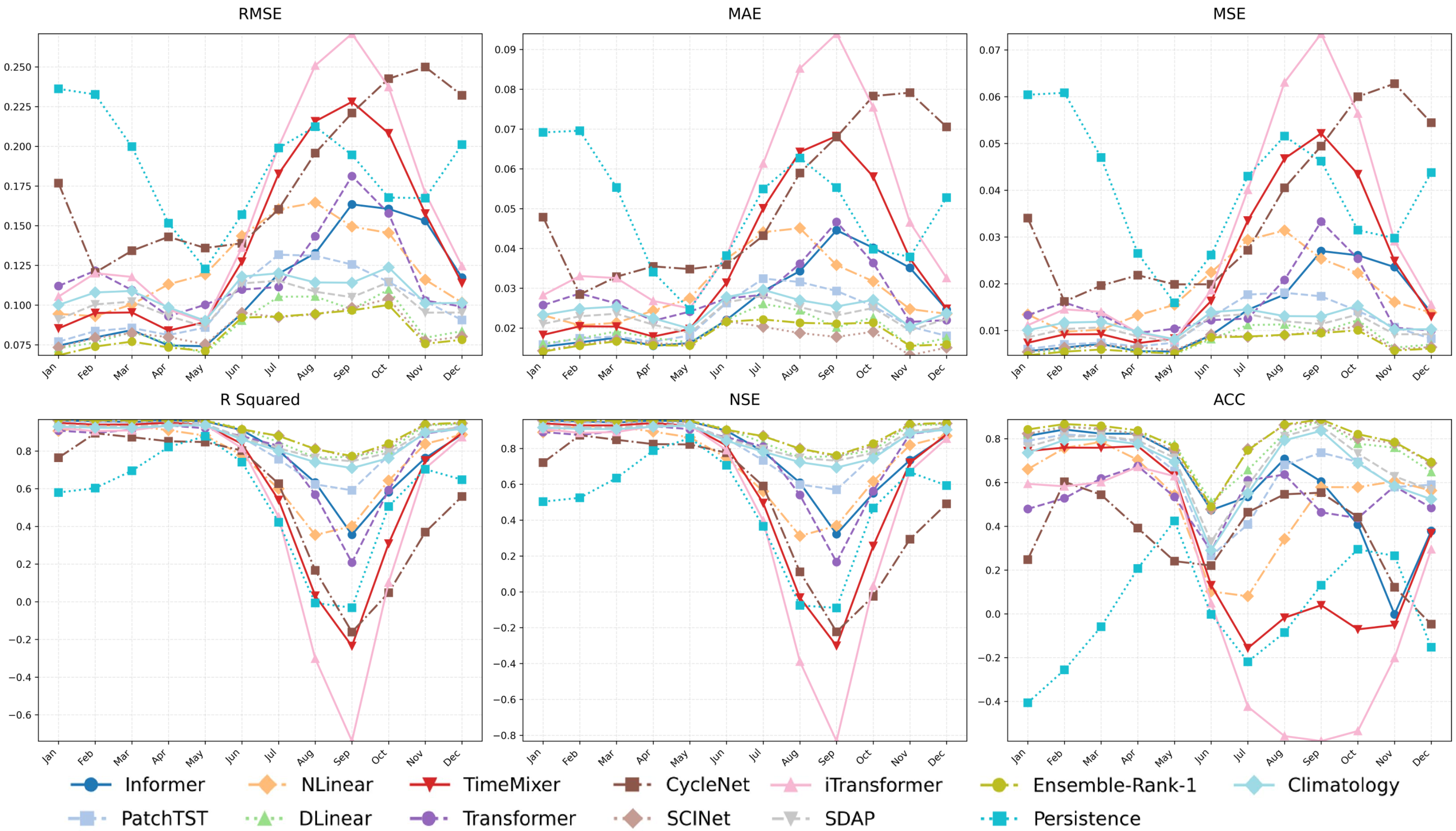}
    \caption{\textbf{S2S forecasting of calendar months over testing period.} During the melting season, when sea ice concentration fluctuates the most, the performance of all models drops abruptly, indicating the unsolved challenge of finding predictive skills throughout the summertime.}
    \label{fig:exp_overall_calendar_month}
\end{figure*}

In this section, we report the primary results and analysis following the S2S forecasting task identified in the previous \textit{Task Overview}\ref{subsec:task} section.
%Section~\ref{subsec:task}.
Moreover, we provide extreme case studies and gain deep insight into the robustness of deep learning approaches.
% \begin{figure}[t]

\subsection{Effectiveness of Latent Space Representation}
% Training on CDR, Test on NT, BT(Heat Map)
To answer \textbf{RQ1}, we pre-train the encoder and decoder on the training dataset corresponding to different retrieval algorithms (CDR, NT, and BT). The test results of fidelity of reconstruction from deep latent space representation are presented in Table~\ref{tab:sic_compress}, which shows satisfactory results on test sets. The average SSIM is around 0.90, MSE is below 0.0033, and NSE is over 0.96. The result shows that the compressed representations can effectively maintain the key spatial features. On average, reducing the data dimensions from 4096 to 1024 causes a marginal 2 - 3\% drop in reconstruction accuracy, which is acceptable considering the computational load it saves.
%While the slight decrease in performance is acceptable, it is because time-series models often lead to lower prediction accuracy as dimensions increase. 
Additionally, to test the robustness of the compressed latent feature, we applied the CDR-trained encoder-decoder to the NT and BT datasets. While MAE saw slight increases, the model still performed well. For example, when applying CDR$_{S}$ to NT, MAE rose from 0.0125 to 0.016. Despite these minor drops, the CDR-trained encoder-decoder demonstrates good robustness by achieving relatively stable results across different datasets. The results of our robustness analysis are presented in Table~\ref{tab:sic_compress_roubustness}.

\subsection{S2S Forecasting Skill}

\textbf{Overall performance.} 
For answering \textbf{RQ2}, we train SIFE in both a non-autoregressive and a 15-day autoregressive roll-out manner. Specifically, in the former setting, the prediction length of SIFE is set to 7 and 6, which correspond to the daily and monthly average scale.
The overall performance of DL models, SDAP, persistence, and climatology baselines under the S2S forecasting scenario is given in Table~\ref{tab:ts_forecasting}.
For a more conventional temporal scale, i.e., forecasting 7 days and 6 months' average of SIC, all DL models showcase promising results in terms of accuracy (MSE, MAE) and predictive skill (ACC, R$^2$). 
The persistence performs the best in short-term forecasting, since it assumes the state of sea ice remains in subsequent days and the variations of SIC are usually limited for a week.
For the S2S forecasting scenario, it becomes more challenging for DL models.
In Figure~\ref{fig:exp_overall}, we find that all DL forecasting backbones except for SCINet and DLinear suffer a severe performance drop as the forecasting lead approaches 180 days. 
Since DL models perform S2S forecasting in an autoregressive manner, the accumulated predictive errors inevitably compromise models' ability to capture the genuine long-term patterns.
Specifically, Figure~\ref{fig:exp_overall_calendar_month} showcases the primary source of error where all models perform the worst, i.e., the melting season. 
During the summer time, when Arctic sea ice varies the most, the loss of predictability of SIC leads to the inferior forecasting skill of DL models. 
However, SCINet, Informer, and DLinear models are less prone to the drop of sea ice predictability, indicating that decomposing subsequent features, modeling of long-term dependencies, and employing a combination of simple linear models could be an important factor to mitigate the challenge posed by S2S daily forecasting.
To further leverage trained SIFE with different time series backbones, we explore the ensemble of different members in Table~\ref{tab:ensemble}. After fine-tuning the ensemble weights of the top-4 performance SIFE, we find that it could further reduce the RMSE and improve the forecasting skill in terms of ACC. Hence, we propose \textbf{SIFE-Ensemble-Rank-1} as the DL baseline for S2S daily Arctic sea ice forecasting.

\noindent\textbf{Sea Ice Outlook Prediction.} To explore the potential of SIFE to mitigate the challenging spring barrier (\textbf{RQ3}), we introduce the SIO into our benchmark. It is a project that collects, analyzes, and synthesizes the best real-time seasonal predictions of September Arctic SIE from various institutes and agencies across the globe. It has been collecting predictions initialized on the first day of June, July, August, and added predictions initialized on September 1, providing necessary sea ice prediction information for socioeconomic activities in the Arctic region. The SIO reports forecasts from 2008 to 2021, and they can be obtained from the online repository. To evaluate the SIE and SIC forecasting skill of the SIFE, we align the forecast starting time of SIO to our test set, and compare their reported forecasts with predictions of both fixed-parameter and fine-tuned SIFE. The detailed results can be found in Table~\ref{tab:sio_forecasting}. The SIO clearly suffers from the SPB; their forecasting skill drops rapidly when the forecast date is earlier than May (120 days lead time), while our SIFE models are more skillful and more stable. Although the near September forecasts are less accurate than the SIO prediction, the consistent forecasting skill could indicate that our SIFE is reliable and less prone to the SPB.

% TBD Surprisingly Good Performance of Climatology

\begin{figure}[t]
    \centering
    \includegraphics[width=0.7\linewidth]{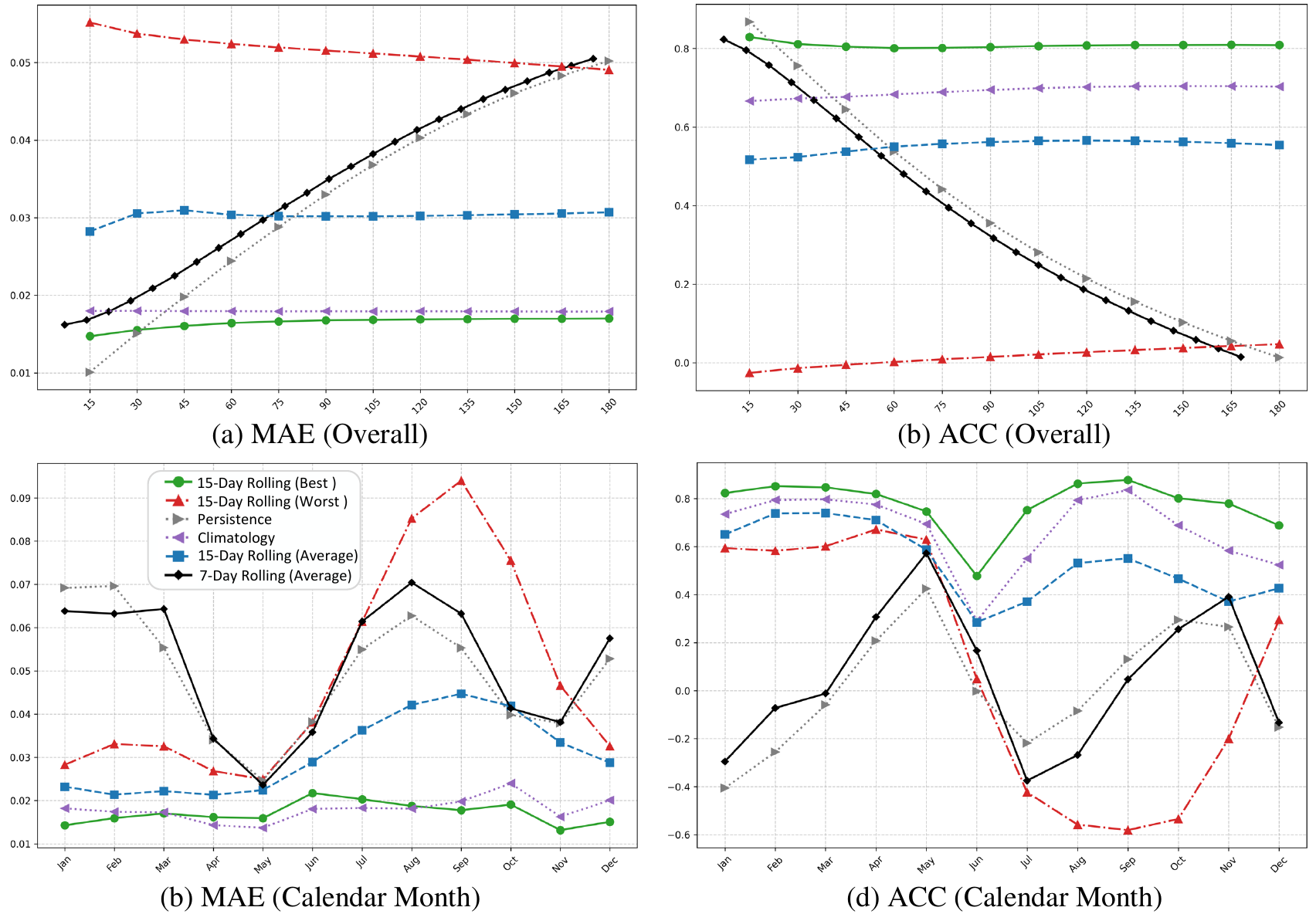}
    % \vspace{-0.5cm}
    \caption{Comparison between S2S forecasting adopting the 7-day rolling approach and the 15-day rolling-based training strategy. The forecasting skill (ACC) of longer rolling windows is more stable than that of shorter ones. }
    \label{fig:exp_7_days_rolling}
\end{figure}

\noindent\textbf{15-day rolling-based strategy.} To verify our strategy for a 15-day rolling window in the training stage,
%rather than directly adapting a 7-day forecasting window for S2S forecasting, 
we compare it with a 7-day rolling window to forecast 180 days and plot the results in Figure~\ref{fig:exp_7_days_rolling}. Although the backbones are identical, the average performance of 7-day forecasting models is consistently inferior to the S2S forecasting models, indicating the effectiveness of our training strategy.

\begin{table}[t]\small
	% \begin{wraptable}{r}{0.6\textwidth}\small
		\centering
		% \vspace{-1cm}
		\caption{Different backbone members in each SIFE-Ensemble rank: SCINet (R1), Informer (R1), DLinear (R1), PatchTST (R1), Transformer (R1-2), NLinear (R1-2), TimeMixer (R1-2), iTansformer (R1-3), CycleNet (R1-3). We propose best performed model ensemble, \textbf{SIFE-Ensemble-Rank-1}, to be the DL baseline.}
		\begin{tabular}{c|c|ccc}
			\toprule
			\textbf{SIFE-Ensembles} & Members   & \textbf{MAE}$\downarrow$ & \textbf{RMSE}$\downarrow$   &  \textbf{ACC}$\downarrow$ \\ \hline
			Rank-1 & Top-4 & \textbf{0.0175} & \textbf{0.0776} & \textbf{0.8155}  \\ 
			
			Rank-2 & Top-7 & 0.0241 & 0.0884 &  0.8100     \\ 
			
			Rank-3 & All & 0.0277 & 0.0946 &  0.7804   \\ 
			 \bottomrule
		\end{tabular}
		%\caption{Different backbone members in each SIFE-Ensemble rank: SCINet (R1), Informer (R1), DLinear (R1), PatchTST (R1), Transformer (R1-2), NLinear (R1-2), TimeMixer (R1-2), iTansformer (R1-3), CycleNet (R1-3). We propose best performed model ensemble, \textbf{SIFE-Ensemble-Rank-1}, to be the DL baseline.}
		\label{tab:ensemble}
        \vspace{-0.1cm}
	\end{table}
	
	% \end{wraptable}

% \begin{wrapfigure}{r}{0.65\columnwidth}
% \vspace{-0.5cm}
\vspace{-0.2cm}
% \end{wrapfigure}

\begin{figure}[t]
	\vspace{-0.5cm}
	\centering
	\includegraphics[width=0.65\linewidth]{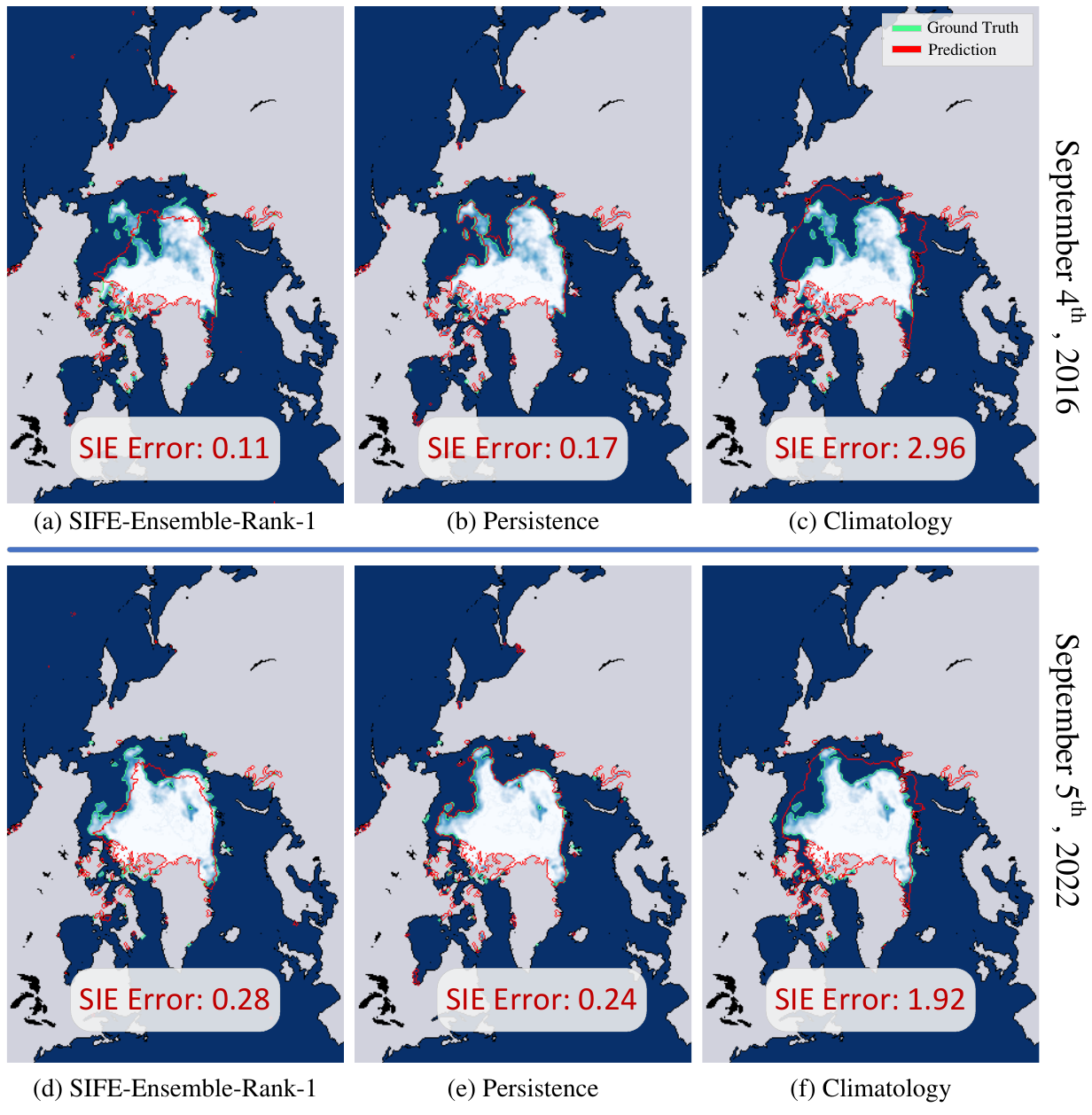}
	\vspace{-0.3cm}
	\caption{Observed and predicted minimum of sea ice extent in September 2016 and 2022.}
	\label{fig:exp_extreme}
	% \end{figure}
% \vspace{-0.3cm}
\end{figure}

% Extreme sea ice extent predicted by the SIFE model, persistence and climatology.

\subsection{Extreme Case Study}
%Annual Extreme in test period
We present annual and decadal minimum SIE
%, as mentioned in Section~\ref{subsec:Methodology},
predicted by the proposed DL baseline, persistence, and climatology in Figure~\ref{fig:exp_extreme}(a)-(f). Although our DL model showcases relatively superior S2S performance, predicting extreme events remains a challenging task. More visualization and residual analysis are attached to the Appendix.
%including hindcast performances, 

% \begin{figure}[t]
%     \centering
%     \includegraphics[width=1.0\linewidth]{Figs/Experiments/exp_fig_ExtremeVisual_2016.pdf}
%     \caption{\textbf{Observed decadal minimum of sea ice extent in September 2016.} Placeholder.}
%     \label{fig:exp_overall}
% \end{figure}

% \begin{figure}[t]
%     \centering
%     \includegraphics[width=1.0\linewidth]{Figs/Experiments/exp_fig_ExtremeVisual_2022.pdf}
%     \caption{\textbf{Observed annual sea ice extent in September 2022.} Placeholder.}
%     \label{fig:exp_overall}
% \end{figure}

% Hindcast: Attach to appendix

% \subsection{Down-scaling.}

\section{Conlusion}
% Propose the first benchmark regarding S2S sea ice forecasting.
We present IceBench-S2S, the first benchmark addressing the challenging and scientifically significant S2S daily sea ice forecasting.
% Showcasing capability of DL models forecasting in latent space.
Following recent advances in climate forecasting, we perform forecasting in a deep latent space by leveraging spatial autoencoders to effectively compress daily sea ice data. Since we aim to extend daily forecasting leads from the commonly studied sub-seasonal scale to 180 days, skillful long-term time-series forecasting models are leveraged as backbones to capture sea ice variation through latent space. Our approach not only bridges the gap between sea ice forecasting and the state-of-the-art time series DL models but also forms a competitive DL baseline and poses a critical challenge for DL models to resolve.
% However, the improved linear-based (Dlinear,Nlinear) models betters most of the models.
The comprehensive experiments showcase that forecasting backbones adopting subsequence decomposition and a combination of linear models could mitigate the loss of predictability during the melting seasons, indicating a potential direction for future study.

% SCINet as backbone has shown potential

\noindent\textbf{Limitation and Future Work.}
% Incoporating more atmospheric and oceanic variables.
%
%the state-of-the-art
Considering the crucial impact posed by the Arctic sea ice, we will incorporate correlated atmospheric and oceanic variables in our future work to further investigate the potential of deep learning models for revealing fundamental atmosphere-sea-ice coupling patterns.

% \bibliography{references}  %%% Uncomment this line and comment out the ``thebibliography'' section below to use the external .bib file (using bibtex) .

\bibliographystyle{unsrtnat}
\bibliography{reference}

\end{document}